\documentclass[english, 10pt, twocolumn, twoside]{IEEEtran}  % twocolumn

\usepackage{threeparttable}
\usepackage{cite}
\ifCLASSINFOpdf
\else
\fi

\usepackage{graphics}
\usepackage{epstopdf}
\usepackage{subfigure}
\usepackage[cmex10]{amsmath}
\usepackage{amsfonts}
\usepackage{cases}
\usepackage{booktabs}
\usepackage{epsfig}
\usepackage{url}
\usepackage{algorithm}
\usepackage{algorithmic}
\usepackage{array}
\usepackage{multirow}
\usepackage{bigstrut}
\usepackage{amssymb, bm}
\usepackage{color}
\usepackage{booktabs}
\usepackage{comment}

\begin{document}

\title{FPCR-Net: Feature Pyramidal Correlation and Residual Reconstruction for Optical Flow Estimation}

\author{Xiaolin~Song,
        Yuyang~Zhao,
        Jingyu~Yang,
        Cuiling~Lan,
        and~Wenjun~Zeng

        % <-this % stops a space

}% <-this

\maketitle

\begin{abstract}
% 11 lines

Optical flow estimation is an important yet challenging problem in the field of video analytics. The features of different semantics levels/layers of a convolutional neural network can provide information of different granularity. To exploit such flexible and comprehensive information, we propose a Feature Pyramidal Correlation and Residual Reconstruction Network (FPCR-Net) for optical flow estimation from frame pairs. It consists of two main modules: pyramid correlation mapping and residual reconstruction. The pyramid correlation mapping module takes advantage of the multi-scale correlations of global/local patches by aggregating features of different scales to form a multi-level cost volume. The residual reconstruction module aims to reconstruct the sub-band high-frequency residuals of finer optical flow in each stage. Based on the pyramid correlation mapping, we further propose a correlation-warping-normalization (CWN) module to efficiently exploit the correlation dependency. Experiment results show that the proposed scheme achieves the state-of-the-art performance,  with improvement by 0.80, 1.15 and 0.10 in terms of average end-point error (AEE) against competing baseline methods --- FlowNet2, LiteFlowNet and PWC-Net on the \emph{Final} pass of Sintel dataset, respectively.
\end{abstract}


\IEEEpeerreviewmaketitle

\section{Introduction}
\label{secIntroduction}

\IEEEPARstart{O}{ptical} flow estimation is an important yet challenging problem in the field of video analytics. Recently, deep learning based approaches have been extensively exploited to estimate optical flow. Despite the great efforts and rapid developments, the advancements are not as significant as those achieved in single image based computer vision tasks. The main reason is that optical flow is not directly measurable in the wild and it is challenging to model motion dynamics with pixel-wise correspondence between two consecutive frames. Thus optical flow estimation requires the deep learning model to train with more samples consisting of different displacements, and most importantly formulated with more effective architecture. To increase the utilization of training data, we propose a novel loss function to incorporate the best of supervision and unsupervision. The supervised training ensures the unsupervised learning with more significant representation, while the unsupervised alignment reduces dependencies of the network on the ground-truth optical flow and helps for quick convergence from a few of training pairs.

Conventional methods attempts to propose mathematical algorithms of optical flow estimation such as DeepFlow \cite{Weinzaepfel2013DeepFlow} and EpicFlow \cite{revaud2015epicflow} by matching features of two frames. However, these methods are complex with high computational complexity, and usually fail for motions with large displacements. Convolutional neural networks (CNNs) like FlowNet \cite{dosovitskiy2015flownet}, SpyNet \cite{ranjan2017optical}, PWC-Net \cite{Sun_2018_CVPR} boost the state-of-the-art performance of optical flow estimation and outperform conventional methods, with effective structures for feature correlation and warping. Most of these models, however, suffer from difficulties in model training and limitations, such as lack of joint feature correlation in each level for fusion representation, or absence of residual learning with representation associated in stages for taking more details of optical flow and compensating object occlusions between frames. Moreover, most of those approaches cannot leverage specialized detail refinement for obtaining further performance enhancement even though more datasets are fine-tuned to refine network for different displacement limited by the network designs.

In this paper, we propose an end-to-end architecture with jointly and effectively feature-wise pyramid correlation representation and residual learning in each stage, which is capable of exploring the granular dynamics of consecutive frames. The proposed pyramid correlation mapping correlates multi-level features and jointly embeds the cost volumes in different scales to yield more edge or texture correlation details. The residual reconstruction module predicts
the residuals in a coarse-to-fine architecture, which is used to refine predicted optical flow for multi-stage optical flow estimation networks. In addition, we present a correlation-warping-normalization (CWN) module involving pyramid features of input frames. After the pyramid representations of multi-level correlation features, it is essential to use a efficient module, \emph{i.e.} CWN module, to deploy the cost volume from pyramid correlation mapping operation, which is a feature-level fusion for multi-correlation learning to estimate flow accurately, especially for small displacements.

To summarize, the main contributions of this work are three-fold:

\begin{itemize}

\item We propose a pyramid correlation mapping operation for jointly embedding the cost volumes in different scales, to yield detailed motion information in multi-scale features for temporal modeling. And we present a CWN module for fusing features from backbone and reconstruction branch, to refine intrinsic flow estimation by optimizing warped feature and cost volume.
\item We propose a residual learning branch for optical flow reconstruction, to further reconstruct the sub-band high-frequency residuals of finer optical flow in each stage, which can better exploit the temporal dynamics to a more accurate and effective representation and occlusion compensation between two frames.
\item We propose a novel loss function for optical flow estimation, which uses supervised as well as unsupervised learning cues.
\end{itemize}

\section{Related Work}
\label{secRelWork}

Horn and Schunck \cite{horn1981determining} pioneer the study on optical flow estimation. Brox \emph{et al.} \cite{brox2004high} take advantage of illumination changes by combining the brightness and propose the warping-based estimation method. Brox \emph{et al.} \cite{brox2010large} aggregate rich descriptors with feature matching into the variational formulation. Weinzaepfel \emph{et al.} \cite{Weinzaepfel2013DeepFlow} propose DeepFlow to correlate multi-scale patches to generate cost volume and incorporate correlation operation as the matching term in functional. Revaud et al. \cite{revaud2015epicflow} propose EpicFlow that uses externally matched flows as initialization and interpolates them to dense flow. Zimmer et al. \cite{zimmer2011optic} propose the anisotropic smoothness term for complementary regularization to the data term.

CNN-based methods achieve a breakthrough on optical flow estimation. some methods use CNN models for image patches matching.

 Dosovitskiy \emph{et al.} \cite{dosovitskiy2015flownet} establish FlowNet which is the important CNN exploration on optical flow estimation with encoder-deconder architecture, of which FlowNetS and FlowNetC are proposed with simple operations. However, the number of parameters is large with heavy calculation on correlation. Ilg \emph{et al.} \cite{Ilg_2017_CVPR} propose a cascaded network with milestone performance based on FlowNetS and FlowNetC with huge parameters and expensive computation complexity.

Ranjan \emph{et al.} \cite{ranjan2017optical} present a compact network named SPyNet from is inspired from spatial pyramid with multi-level representation learning. Nevertheless, the performance is not significant. To reduce the number of parameters, Hui \emph{et al.} \cite{hui2018liteflownet} leverage a compact LiteFlowNet and Sun \emph{et al.} \cite{Sun_2018_CVPR} propose PWC-Net with high accuracy for optical flow estimation, which are pioneers of the trend to lightweight optical flow estimation networks. They utilize light feature-level matching and warping motivated by conventional methods. LiteFlowNet \cite{hui2018liteflownet} involves cascaded flow inference for flow warping and feature matching, and feature-driven local convolution (f-lconv) for flow regularization. PWC-Net \emph{et al.} \cite{Sun_2018_CVPR} utilizes feature pyramid extraction and feature warping to construct the cost volume, and uses context network for optical flow refinement. Our method introduces an unsupervised term, a flow regularization term for loss function and a global refinement branch to learn reconstruction residuals in each stage for accurate flow estimation inspired by conventional methods and CNN-based methods.

\section{FPCR-Net}
\begin{figure*}[t]
 \centering\includegraphics[width=0.9\textwidth]{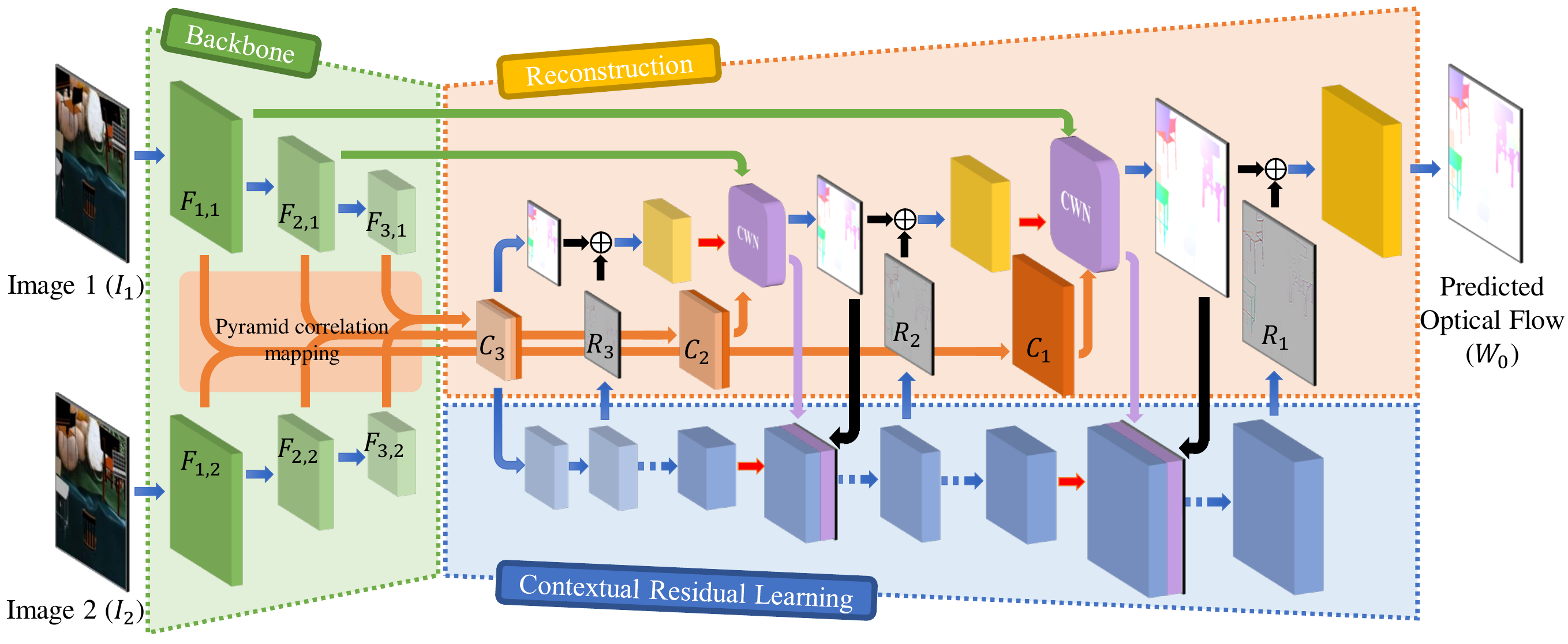}
 %\vspace{-2mm}
 \caption{The overall framework with our Feature Pyramid Correlation and Residual Learning Network (FPCR-Net) for optical flow estimation (only a 3-stage design is shown). The blue and red arrows are convolution and transposed convolution operations, respectively. The reconstruction branch yields the correlation features $\mathcal{C}_k$ from the backbone which are optimized by the CWN module, and the generated optical flow is refined by residual features $R_k$ from the residual learning branch.}
 \label{fig:arch}
% \vspace{-3mm}
\end{figure*}

Optical flow denotes the dense motions between two consecutive frames, and optical flow estimation is the process of modeling temporal dynamic correspondence densely. The features of different semantics levels/stages of a convolutional neural network can provide representations of different granularity and compensation. Figure~\ref{fig:arch} shows the overall flowchart of our overall framework called feature pyramid correlation mapping and residual learning network for optical flow estimation, \emph{i.e.} FPCR-Net.
 To exploit such flexible and comprehensive information for motion dynamics, we take the main reconstruction branch modified from FlowNetC \cite{dosovitskiy2015flownet} with the pyramid correlation mapping and CWN module embedding, and the residual learning branch utilizing for multi-stage fine-grained feature reconstruction.

\subsection{Pyramid Correlation Mapping}

\begin{figure}[t]
\begin{center}
\includegraphics[width=0.95\linewidth]{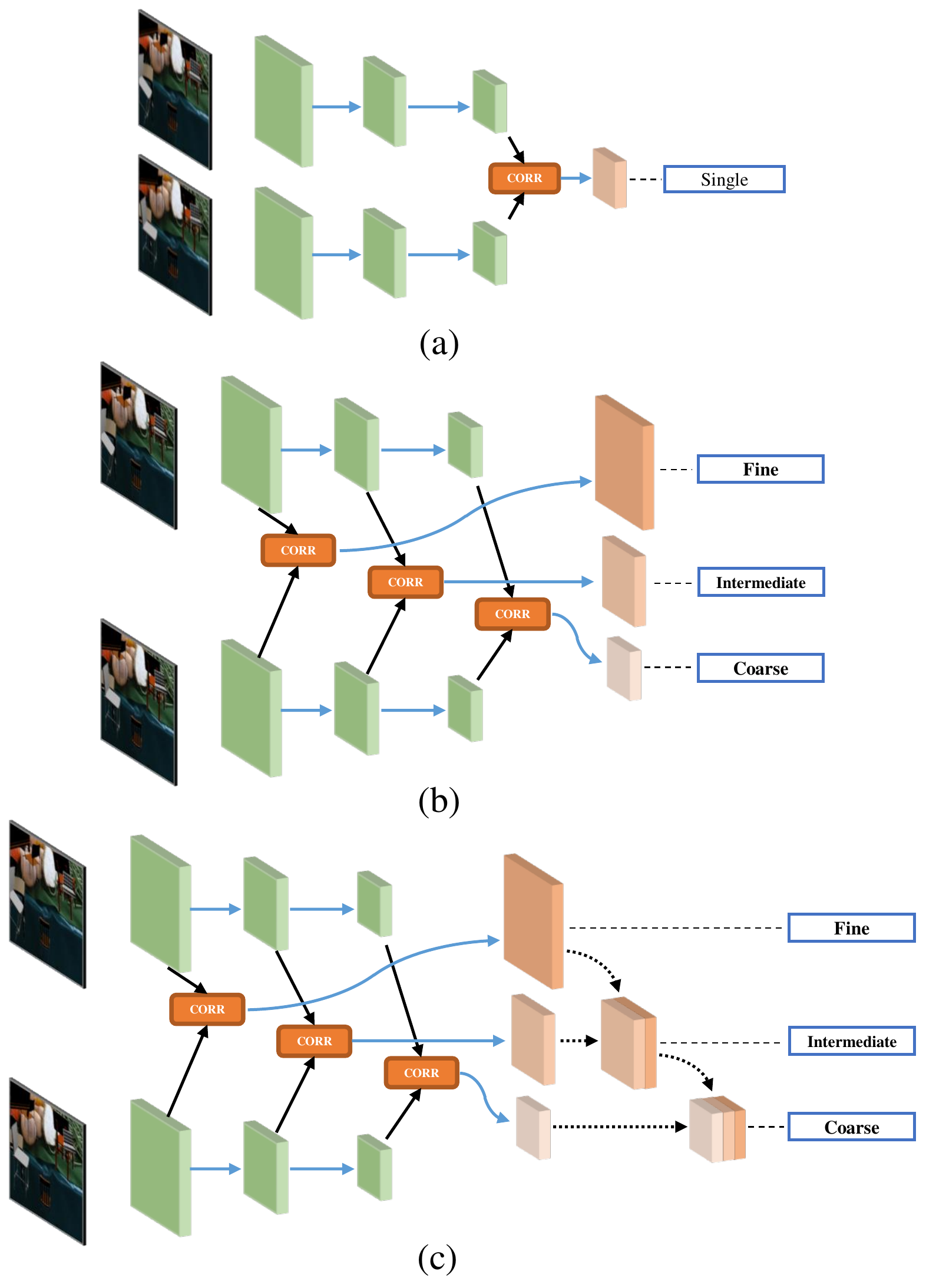}
\end{center}
%\vspace{-7mm}
\begin{center}
\caption[width=0.95\linewidth]{Illustration of the pyramid correlation mapping operation, which calculates and aggregates multi-scale cost volumes. (a) single correlation of high-level low-resolution features, (b) correlation of multi-level features, (c) correlation and mapping of multi-level features.}
\label{fig:corr}
\end{center}
%\vspace{-5mm}
\end{figure}

%\ref{figcorr}
Most CNN-based methods for optical flow estimation use frame feature extractor (backbone) modified from FlowNetC \cite{dosovitskiy2015flownet} with down-sampling and feature-level correlation of input frames. Each layer extracts corresponding features to transform the two input frames to pyramidal multi-scale and multi-dimensional representation with shared weights. To reduce the computational complexity and take advantage of multi-level features, inspired by LitFlowNet \cite{hui2018liteflownet}, we perform short-range correlation for each pyramid level features, instead of long-range correlation at a single level like FlowNetC. Moreover, to reduce the burden of correlation, we use an atrous convolution pyramid module with stride in feature extraction for sparse correlation instead of traditional convolution. Besides, we propose a pyramid correlation mapping operation for multi-level cost volume, to yield more edge or texture correlation details for different scales. Figure \ref{fig:corr} shows the process. For each frame of an input pair, atrous convolutions with the dilation rates of 1, 2, 4, and 8 are used for extracting feature maps at each convolutional layer to expand the correlation search region and preserve the details of edges and textures. 

Let $F_{k,i}$ denote the $k$-th level feature extracted from the network of input image $I_i$. Specially, $F_{0,i}$ means the original image $I_i$. The single correlation at the $k$-th level is calculated as follows.

\begin{equation}
\begin{split}
\mathcal{C}_{k}^s (\mathbf{x}_1,\mathbf{x}_2)= \sum\limits_{\mathbf{o}} {(\mathbf{f}_{k,1} (\mathbf{x}_1+\mathbf{o}))^\top \mathbf{f}_{k,2} (\mathbf{x}_2+\mathbf{o})},
\end{split}
\end{equation}

\noindent where $\mathbf{o}$ denotes the offset of correlation operation, and $\mathbf{o} \in [-n,n] \times [-n,n]$ for search region. $\mathbf{f}_k$ is the flattened column vector of $F_k$.
We leverage a pyramid correlation mapping operation based on single correlation for aggregating different level cost volumes. The details of this calculation are shown as:

\begin{equation}
\begin{split}
\mathcal{C}_{k} =
\begin{cases}
  \mathcal{C}_{1}^s, & \mbox{$k=1$}, \\
  \mathcal{C}_{k}^s \odot (\mathcal{C}_{k-1} \Downarrow), & \mbox{$k>1$},
\end{cases}
\end{split}
\end{equation}

\noindent where $\Downarrow$ is the down-sampling operator with average pooling and channel reduction at the rate of $\frac{1}{2}$, and $\odot$ denotes the concatenating across the channel dimension.

\subsection{Correlation-Warping-Normalization Module}
\label{sseccwn}

\begin{figure*}[t]
\begin{center}
\includegraphics[width=1.0\linewidth]{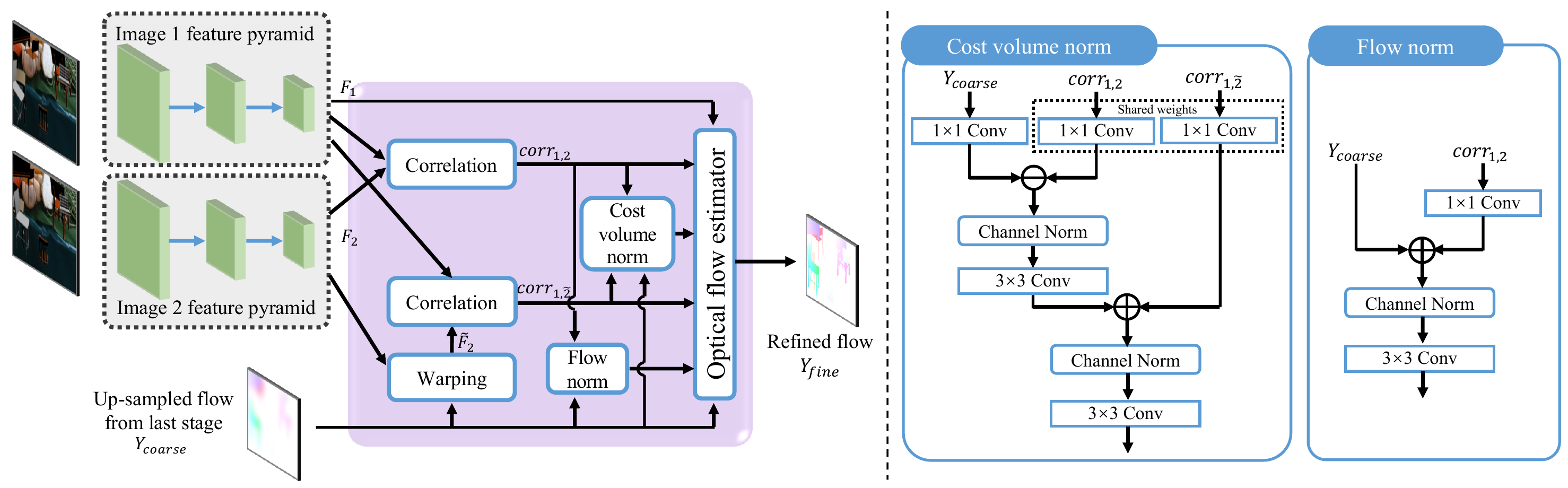}
\end{center}
%\vspace{-7mm}
\begin{center}
\caption[width=0.95\linewidth]{Details of CWN module with correlation, warping and normalization units.}
\label{fig:cwn}
\end{center}
%\end{minipage}
%\vspace{-5mm}
\end{figure*}

Inspired by FlowNet2 \cite{Ilg_2017_CVPR} and PWC-Net \cite{Sun_2018_CVPR}, we propose the multi-component warping and correlation module within the reconstruction branch, \emph{i.e.} Correlation-Warping-Normalization (CWN) Module, shown in Figure \ref{fig:cwn}. This module fuses the cost volumes from pyramid correlation mapping and involves the normalization operator for cost volumes to feed into multi-layer CNN flow estimator, for refining the up-sampled optical flow with both large and small displacement motion modeling.

\noindent\textbf{Correlation and Warping.} For stage $k$ of the reconstruction, the features $F_{k,i}$ construct a cost volume which contain the patch-wise matching scores by pyramid correlation mapping $\mathcal{C}_k$. In addition, the feature map $F_{k,2}$ is warped to the view of $F_{k,1}$, denoted by $\tilde F_{k,2}$, via the bicubic up-sampling flow from stage $k-1$. Then $\tilde F_{k,2}$ is correlated with $F_{k,1}$, and the cost volume contains errors from the coarse features, especially around the edges on $I_1$, indicating the implicit optimization attention region for auxiliary learning of $\mathcal{C}_k$. Besides, the motion range of this operation is set to a small value of 4, further reducing parameters in correlation mapping.

\noindent\textbf{Cost Volume Normalization.} Compared with other normalization methods, channel-wise normalization is much more useful in this task to aid generalization for feature representation. We propose the channel normalization operation, and establish the cost volume normalization and flow normalization for correlation feature fusion and normalization. Denote the feature as $V$, and the channel normalization defines as:

\begin{equation}
\begin{split}
\mathcal{N}(\mathbf{x},c) = {{V(\mathbf{x},c)} \mathord{\left/
 {\vphantom {{V(\mathbf{x},c)} {\left( {\alpha \sum\limits_\mathbf{x} {\left( {V(\mathbf{x},c)} \right)^2 + \epsilon} } \right)^\beta}}} \right.
 \kern-\nulldelimiterspace} {\left( {\alpha \sum\limits_\mathbf{x} {\left( {V(\mathbf{x},c)} \right)^2 + \epsilon} } \right)^\beta}} ,
\end{split}
\end{equation}

\noindent where $c$ denotes the channel of $V$. $\alpha$, $\beta$ and $\epsilon$ denote the multiplier, the exponent and the additive constant with a small value for normalization term, respectively. We use $\alpha = 0.99$, $\beta = 0.5$ and $\epsilon = 0.01$. %This operator is suitable for optical flow $W (\mathbf{x})$, and the output is the relative motion amplitude value of $u$ and $v$ components in pixel.

\subsection{Residual Reconstruction}

\begin{figure}[t]
%\begin{minipage}[t]{0.49\linewidth}
\begin{center}
\includegraphics[width=0.8\linewidth]{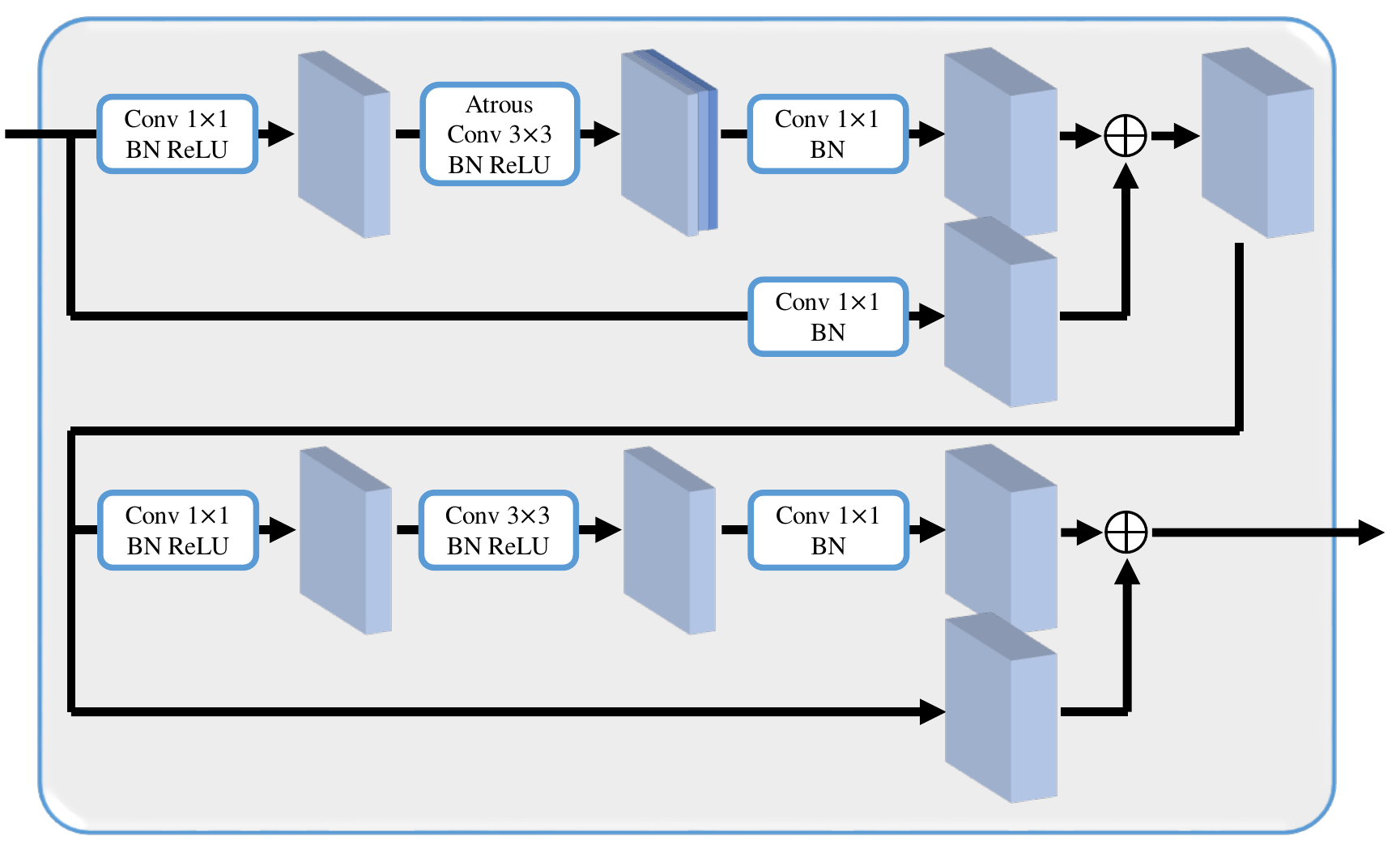}
\end{center}
%\vspace{-7mm}
\begin{center}
\caption[width=0.95\linewidth]{Convolution unit in the residual learning branch, which is stacked by different output channels for refinement.}
\label{fig:refine_conv}
\end{center}
%\vspace{-5mm}
\end{figure}

\begin{figure}[t]%[p]
	\begin{center}
		\includegraphics[width=0.95\linewidth]{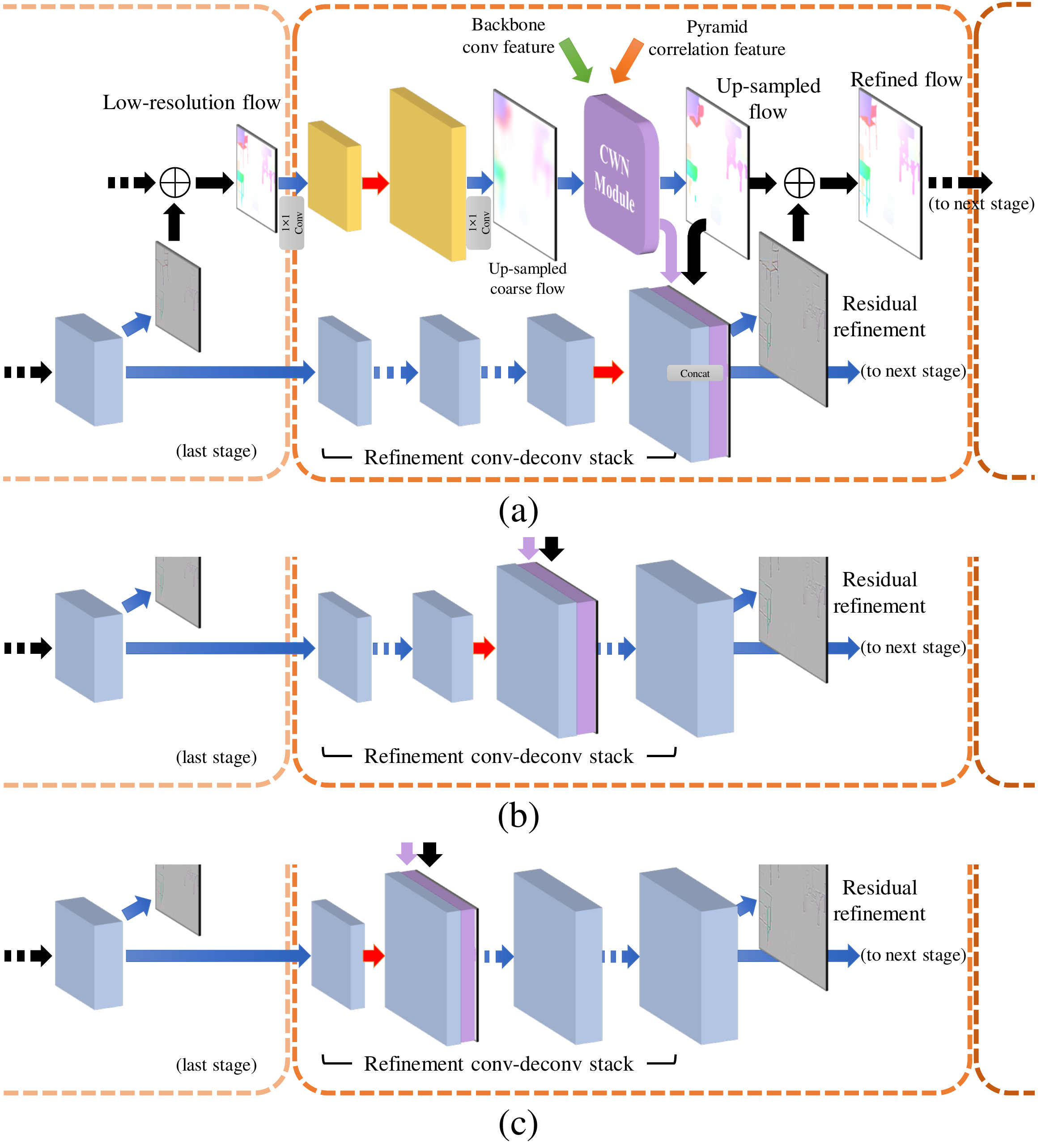}
		%\vspace{-5mm}
		\caption{Details of feature residual learning branch. We compare three types of refinement operations with different up-sampling location. (a) late up-sampling; (b) middle up-sampling; (c) early up-sampling. The blue arrows are the stacks of convolution unit in Figure \ref{fig:refine_conv}, and the red arrows are transposed convolution operation in this branch.}
        \label{fig:refinement}
	\end{center}
%\vspace{-6mm}
\end{figure}

To explore motion details and occlusion compensation between frames to learn fine-grained residual representation, we construct our network by utilizing the pyramid residual learning framework for coarse-to-fine residual learning, as shown in Figure \ref{fig:arch} and Figure \ref{fig:refinement}. For stage $k$, the residual learning branch is independently structured with refining the residual map from stage $k-1$. This branch consists cascades of convolution stack module shown in Figure \ref{fig:refine_conv} for fining residual features and one transposed convolutional layer to up-sample the features by a scale of 2 and fusing with the output of CWN module. The convolution stack module is a cascade of two parts similar to residual blocks in ResNet \cite{he2016deep} --- the atrous projection block and the single residual block. There is no activation function in the last convolution layer of the two block, and the atrous projection block utilizes atrous pyramid convolution at the first layer with the dilated rate of 1, 2 and 4. Due to the similarity of function in each stage, the parameters of the convolution stack module are partially shared except the atrous convolution and the last convolution layer, in order to reduce the parameter number and increase the non-linearity of the network. %Besides, the refinement can learn the occlusion

\subsection{Training loss function}

\begin{figure}[t]
%\begin{minipage}[t]{0.49\linewidth}
\begin{center}
\includegraphics[width=\linewidth]{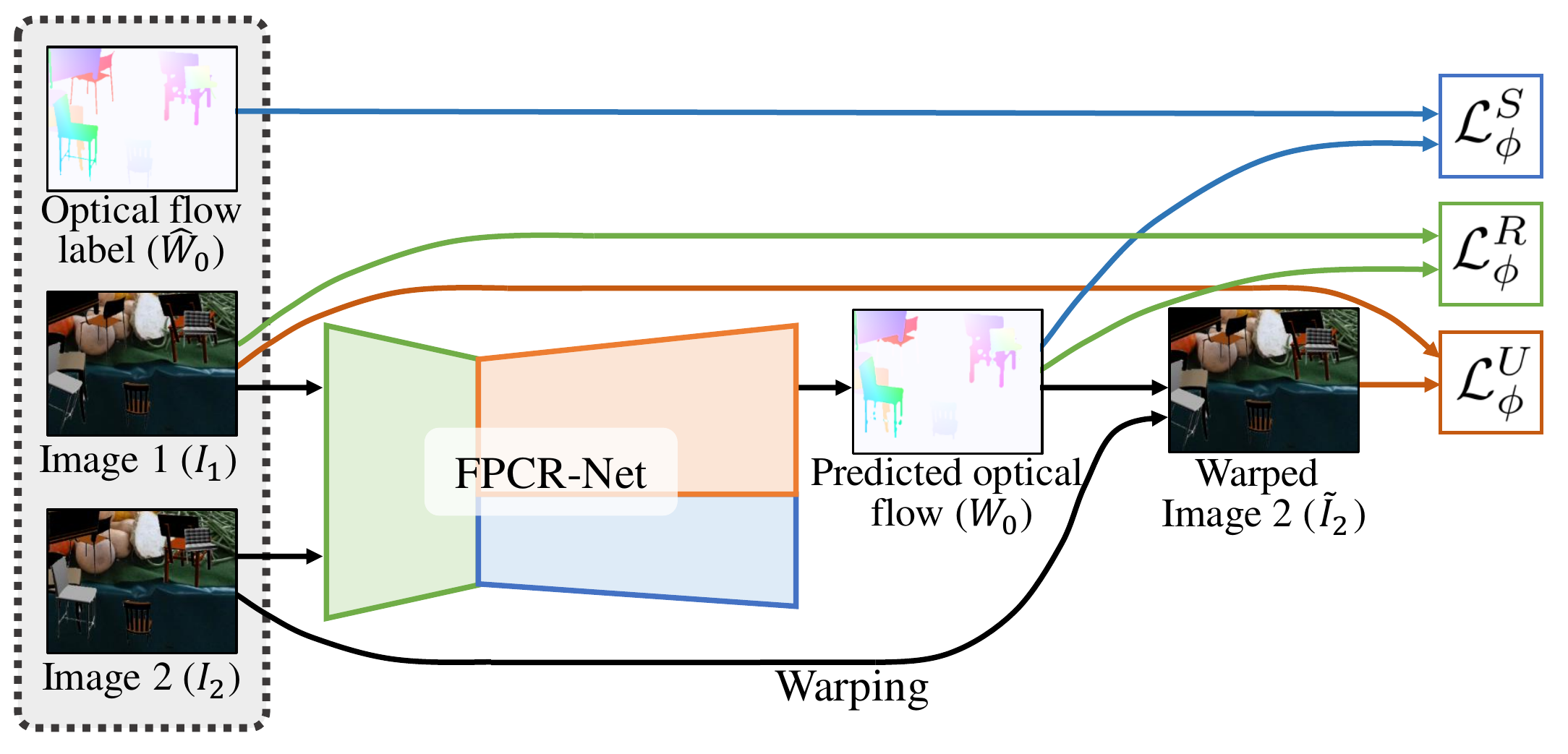}
\end{center}
%\vspace{-7mm}
\begin{center}
\caption[width=0.95\linewidth]{Illustration of the loss function with multiple losses.}
\label{fig:loss}
\end{center}
%\vspace{-5mm}
%\end{minipage}%
\end{figure}

According to the optical flow characteristics, we proposed a loss function incorporating both supervised and unsupervised constraints. For the $k$-th level, the total loss function is as follows.

\begin{equation}
\begin{split}
\mathcal{L}_{\phi,k}  (F_{k,1},F_{k,2},\hat W_k)= \mathcal{L}_{\phi,k} ^S (F_{k,1},F_{k,2},\hat W_k)& \\
+ \lambda \mathcal{L}_{\phi,k} ^U (F_{k,1},F_{k,2}) + \mu \mathcal{L}_{\phi,k} ^R (I_1,I_2)&, \\
\end{split}
\label{eq:loss}
\end{equation}

\noindent where $\phi$ denote the network parameters that predict the optical flow, and $\mathcal{L}_{\phi,k} ^S$, $\mathcal{L}_{\phi,k} ^U$ and $\mathcal{L}_{\phi,k} ^R$ denote the supervised, unsupervised and regularization loss, respectively. Figure \ref{fig:loss} shows the training loss on the original input images $I_i$ (or $F_{0,i}$), ground truth $\hat W_0$ and corresponding predicted optical flow $W_0$. Actually, the loss is used for multi-stage optical flow prediction with different loss weights. $\lambda$ and $\mu$ are the coefficients to balance the three loss term.

\noindent\textbf{Supervised Loss.} The supervised loss term measures the pixel-wise deviation between the predicted optical flow from two frames and the ground truth, and we utilize the end-point error loss as the supervised loss function according to the standard metric on the test of optical flow estimation results. In addition, the smooth-L1 loss is used for spatial constraint optimization.

%\vspace{-3mm}
\begin{equation}
\begin{split}
\mathcal{L}_{\phi,k} ^S = &\sum\limits_\mathbf{x} {\left( {\sum\limits_{d \in \{ u,v\} } {(W_k (\mathbf{x},d) - \hat W_k (\mathbf{x},d))^2 } } \right) } ^ {\frac{1}{2}}  \\
 &+ \sum\limits_{d \in \{ u,v\} } {\sum\limits_\mathbf{x} {\left| {W_k (\mathbf{x},d) - \hat W_k (\mathbf{x},d)} \right| } },
\end{split}
\end{equation}

\noindent where $d$ denotes the $u$ and $v$ direction components of optical flow $W_k$ and $\hat W_k$.  $\circ$ denotes the Hadamard product operator. %Specially, some optical flow labels are sparse sampling in some datasets, such as KITTI dataset. We exclude the invalid pixels or patches in calculate the supervised loss with a mask $M_k$. %In addition, we upsample the predicted flow at the quarter resolution to compare with the scaled ground truth at the full resolution.
\begin{comment}
\begin{equation}
\label{eq:loss}
\begin{split}
\mathcal{L}_{\phi,k} ^S = &\sum\limits_\mathbf{x} {\sqrt {\sum\limits_{d \in \{ u,v\} } {(\mathbf{M}_k \circ (W_k (\mathbf{x},d) - \hat W_k (\mathbf{x},d)))^2 } } }  \\
 &+ \sum\limits_{d \in \{ u,v\} } {\sum\limits_\mathbf{x} {\left| {\mathbf{M}_k \circ (W_k (\mathbf{x},d) - \hat W_k (\mathbf{x},d))} \right|, } }
\end{split}
\end{equation}

\noindent where $d$ denotes the $u$ and $v$ direction components of optical flow $W_k$ and $\hat W_k$. $\mathbf{M}_k$ is the mask with the same shape of $W_k (\mathbf{x})$, which indicates the value of ground truth existing or not filling with 1 or 0, respectively. $W_k (\mathbf{x})$ is the all-one map when the optical flow values are densely filled with the full resolution. In contrary, some optical flow labels are sparse sampling or have patch occlusion in some datasets, such as KITTI dataset. We exclude the invalid pixels or patches in calculate the supervised loss corresponding to the mask map $\mathbf{M}_k$. $\circ$ denotes the Hadamard product operator.
\end{comment}

\noindent\textbf{Unsupervised Loss.} The unsupervised term of loss function optimizes the direct image or feature alignment error according to data fidelity as follows. % in both directions % quantifies

\begin{equation}
\begin{split}
\mathcal{L}_{\phi,k} ^U = \sum\limits_\mathbf{x} {\left| {F_{k,1} (\mathbf{x}) - F_{k,2} (\omega (\mathbf{x},W_k (\mathbf{x})))} \right|},
\end{split}
\end{equation}

\noindent where $\omega (\mathbf{x},W_k (\mathbf{x}))$ is the warping function to generate rectified images or features from $F_{k,2}$ to the view of $F_{k,1}$ by predicted optical flow $W_k (\mathbf{x})$. We evaluate the multi-level alignment loss at the values of $F_{k,1}$ and the reconstructed features of $F_{k,2}$ that warp to a valid location into the second image using linear interpolation for subpixel-level warping due to occlusion or overlapping.

\noindent\textbf{Regularization Loss.} The regularization term penalizes optical flow changes at pixels with low-level intensity variation of $F_{k,1}$, to preserve the smoothness of patches and allow for discontinuities in optical flow map at obvious edges in  $F_{k,1}$, which is shown as:
\begin{equation}
\begin{split}
\mathcal{L}_{\phi,k} ^R = \sum\limits_{d \in \{ u,v\} } \sum\limits_\mathbf{x} {\left| {\nabla W_k (\mathbf{x},d)} \right| \circ \exp (- \left| \nabla F_{k,1} (\mathbf{x}) \right| ) }.
\end{split}
\end{equation}

\section{Experiments}
\label{secExperiments}

In this section, we first describe the training and test datasets, and implementation details. Then we study the effects of different factors in our designed network. Finally, we compare our approach with several state-of-the-art approaches.
%We validate the effectiveness of the proposed framework on two benchmark datasets --- Sintel and KITTI. We first describe the datasets and implementation details. Then we study the effects of different factors in our network. Finally, we compare our approach with many state-of-the-art approaches.

\subsection{Datasets}

FPCR-Net is trained on the FlyingChairs dataset \cite{dosovitskiy2015flownet} and the FlyingThings3D dataset \cite{Mayer2016A}. The FlyingChairs dataset contains about 22k image pairs and their optical flow of chairs on different background images with only planar motions (translation and rotation). The FlyingThings3D dataset \cite{Mayer2016A} is a 3D-version motion object dataset which consists of 22k random scenes with 3D models moving in front of static 3D background scenes. %It indicates the affine transformation with variety motion and lighting changes.  And we validate the effectiveness of the proposed framework on the Sintel dataset \cite{Butler2012A}, which is a popular optical flow estimation benchmark dataset. The MPI Sintel dataset is a large dataset with dense ground truth for both small and large displacement magnitudes. There are 1,041 training image pairs for two passes --- \emph{Clean} and \emph{Final}.

We validate the effectiveness of the proposed framework on the Sintel dataset \cite{Butler2012A}, which is a popular optical flow estimation benchmark dataset. The MPI Sintel dataset is a large dataset with dense ground truth for both small and large displacement magnitudes. There are 1,041 training image pairs for two passes --- \emph{Clean} and \emph{Final}. The \emph{Final} pass contains motion blur and atmospheric effects, while the \emph{Clean} pass has clear edges and single light environment. The labels of optical flow are acquired from rendered artificial scenes with special attention to realistic image properties. % The KITTI 2012 dataset \cite{Geiger2012Are} and the KITTI 2015 dataset \cite{Menze2015Object} contains 394 training pairs totally with large displacements. The ground truth is sparse obtained from real world scenes by capturing the scenes by the RGB camera and LiDAR - a 3D laser scanner simultaneously. Limited to the devices, the labels are sparse and captured from only street scene.
%We conduct experiments on two popular optical flow estimation datasets, namely Sintel and KITTI \cite{Geiger2012Are,Menze2015Object}. The MPI Sintel dataset is a large dataset with dense ground truth for both small and large displacement magnitudes. There are 1,041 training image pairs for two passes - \emph{Clean} and \emph{Final}. The \emph{Final} pass contains motion blur and atmospheric effects, while the \emph{Clean} pass is with clear edges and singular light environment. The labels of optical flow are acquired from rendered artificial scenes with special attention to realistic image properties. The KITTI 2012 dataset \cite{Geiger2012Are} and the KITTI 2015 dataset \cite{Menze2015Object} contains 394 training pairs totally with large displacements. The ground truth is sparse obtained from real world scenes by capturing the scenes by the RGB camera and LiDAR - a 3D laser scanner simultaneously. Limited to the devices, the labels are sparse and captured from only street scene.

\subsection{Data Augmentation}

We use the common augmentations including geometric transformations (translation, rotation and scaling), dynamic changes in low-level attributes (brightness, contrast, gamma and color) and Gaussian noise injection. In addition, we adapt two special schemes for data augmentation.

\noindent\textbf{Motionless Injection.} Considering the dynamic range of predicted optical flow and the complexity of the network, we managed to train our proposed network by mixing motionless frames with all-zero optical flow maps, to reduce the jitter of the predicted flow. We select 3k training images randomly from both FlyingChairs dataset and FlyingThings3D dataset, and each image is filled into both the first and the second frame. Naturally, the motion between the two frames is zero and the optical flow map is all-zero filling.

\noindent\textbf{Motion Reversal.} To improve the utilization of the training data, we take advantage of the dataset containing the optical flow labels referred to both the last frame (forward flow) and the next frame (backward flow), \emph{i.e.} FlyingThings3D dataset. Generally, the frame pair and the forward optical flow are used as the training data. Besides, we reverse the frame pairs with the opposite number of the backward optical flow as the augmented training data.

\subsection{Implementation Details}

\noindent\textbf{Training from Scratch.} According to the training details of FlowNet2, we use the same loss weights among stages with 0.32, 0.08, 0.02, 0.01, 0.005. In each stage, the trade-off weights $\lambda$ and $\eta$ of loss function $\mathcal{L}_{\phi,k}$ are set to 0.05 and 0.005, respectively. In terms of the network architecture, we train our network by the following steps: 1) The backbone and the main branch of reconstruction are trained firstly for 120k epochs using the learning rate schedule $S_{long}$ by motionless initialization, and then using schedule  $S_{long}$ for 150k epochs on the FlyingChairs dataset, and fine-tuning on the FlyingThings3D dataset by the $S_{fine}$ learning rate schedule for 125k epochs introduced in \cite{Ilg_2017_CVPR} \footnote{$S_{long}$ schedule starts from learning rate of $10^{-4}$ and reduces the learning rate by 0.5 at 0.4M, 0.6M, 0.8M, and 1M iterations, and learning rate of $S_{fine}$ schedule starts at 1.2M iterations from $10^{-5}$ and reduces at 1.4M, 1.5M, 1.6M iterations.}. 2) The residual learning branch is trained after the last step with fixed parameters of the backbone and the main branch for 100k epochs using the same schedule of step 1. 3) The whole network is trained without fixed parameters followed by step 2 using the $S_{fine}$ learning rate schedule on the FlyingThings3D dataset. We scale the ground truth flow by 20 for easy training and down-sample it as the supervision labels at different levels.
%The weights of each reconstruction stage $k$ in the training loss are set to be ${0.32, 0.08, 0.02, 0.01, 0.005}$. I

\noindent\textbf{Fine-tuning.} The object and motion are not comprehensive in each task and it is not quite sufficient to transfer the model to another dataset with different scene and movable objects. We fine-tune the networks on the target datasets for better performance, \emph{e.g.} the Sintel dataset, with the learning rate schedule $S_{long}$ for 100k epochs. %and the KITTI dataset

\subsection{Ablation Study}

In this subsection, we will analyze the effectiveness of the proposed pyramid correlation mapping component and the CWN module, and discuss the architecture design of residual learning, respectively.

\noindent\textbf{Effectiveness of Pyramid Correlation Mapping.} Aggregation of multi-level correlated cost volume provides the opportunity to jointly explore the motion details at the video level. We show the performance of our scheme in comparison with the baseline scheme on the Sintel training \emph{Clean} dataset in Table \ref{tab:corr}.  We can see that the ``pyramid correlation mapping'' scheme, with multi-level respective fields, achieves 0.20 and 0.32 decrease in average end-point error (AEE) compared with baseline measured in Sintel training \emph{clean} trained by the FlyingChairs dataset and fine-tuned by the FlyingThings3D dataset. %``Single corr.'', ``Pyramid corr.'', and ``Pyramid corr. mapping'' correspond to the three schemes in Figure \ref{fig:corr}. We can see that ``Pyramid corr. mapping'' scheme achieves 0.20 and 0.32 decrease in average end-point error (AEE) measured in Sintel training \emph{clean} trained by the FlyingChairs dataset and fine-tuned by the FlyingThings3D dataset.

\begin{table}[t]
 \caption{Comparison of training pyramid correlation mapping with different correlation schemes in Figure \ref{fig:corr}. Numbers indicate the AEE on Sintel training \emph{Clean}. ``sc.'' ``pc.'' and ``pcm.'' denotes the three correlation schemes --- single correlation, pyramid correlation and pyramid correlation mapping. The network is trained on the FlyingChairs dataset first and fine-tuning on the FlyingThings3D dataset.}% Pyramid correlation mapping performs better than the other correlation schemes.}
 \begin{center}
  \label{tab:corr}
   \fontsize{9pt}{11pt}\selectfont\centering
   \begin{tabular}{|p{2.2cm}|p{1.4cm}<{\centering}|p{1.4cm}<{\centering}|p{1.4cm}<{\centering}|} %{|p{2cm}|p{1.3cm}|p{1.5cm}|p{1.9cm}|} %{|l|c|c|c|}
     \hline
     Dataset & sc. & pc. & pcm. \\
%             &              &               & mapping \\
     \hline\hline
     FlyingChairs & 3.82 & 3.67 & \textbf{3.62} \\
     FlyingThings3D & 3.11 & 2.88 & \textbf{2.79} \\
     \hline
   \end{tabular}
 \end{center}
%\vspace{-5mm}
\end{table}

\begin{table}[t]
 \caption{Comparison of training CWN with different components on AEE on Sintel training \emph{Clean}. ``C-W'' denotes the correlation and warping module and ``C-W-N'' denotes the entire CWN module.}
 \begin{center}
  \label{tab:cwn}
   \fontsize{9pt}{11pt}\selectfont\centering
   \begin{tabular}{|p{2cm}|p{1.1cm}<{\centering}|p{1.1cm}<{\centering}|p{1.1cm}<{\centering}|p{1.1cm}<{\centering}|}%{|p{2cm}|p{1.3cm}|p{1.5cm}|p{1.9cm}|}  %{|l|c|c|c|}
     \hline
     Dataset & C-W & C-W-N & C-W-N & C-W-N \\
     ~ & ~ & w. sc. & w. pc. & w. pcm. \\
     \hline\hline
     FlyingChairs & 3.54 & 3.51 & 3.45 & \textbf{3.42} \\
     FlyingThings3D & 2.91 & 2.87 & 2.79 & \textbf{2.67} \\
     \hline
   \end{tabular}
 \end{center}
%\vspace{-5mm}
\end{table}

\begin{table}[t]
 \caption{Comparison of residual learning branch with different architectures in Figure \ref{fig:refinement} on Sintel training \emph{Clean}. the early-upsampling scheme performs better than the others with FlyingThings3D fine-tuned.}
 %\vspace{-3mm}
 \begin{center}
  \label{tab:refine}
   \fontsize{9pt}{11pt}\selectfont\centering
   \begin{tabular}{|p{2cm}|p{1.59cm}<{\centering}|p{1.59cm}<{\centering}|p{1.59cm}<{\centering}|} %{|p{2cm}|p{1.3cm}|p{1.5cm}|p{1.9cm}|} %{|l|c|c|c|}
     \hline
     Dataset & late & mid. & early \\
     ~ & up-sampling  & up-sampling  & up-sampling \\
     \hline\hline
     FlyingChairs & 3.58 & \textbf{3.49} &3.50 \\
     FlyingThings3D & 2.87 & 2.76 & \textbf{2.69} \\
     \hline
   \end{tabular}
 \end{center}
%\vspace{-6mm}
\end{table}

\begin{figure*}[t]
	\begin{center}
		\includegraphics[width=\linewidth]{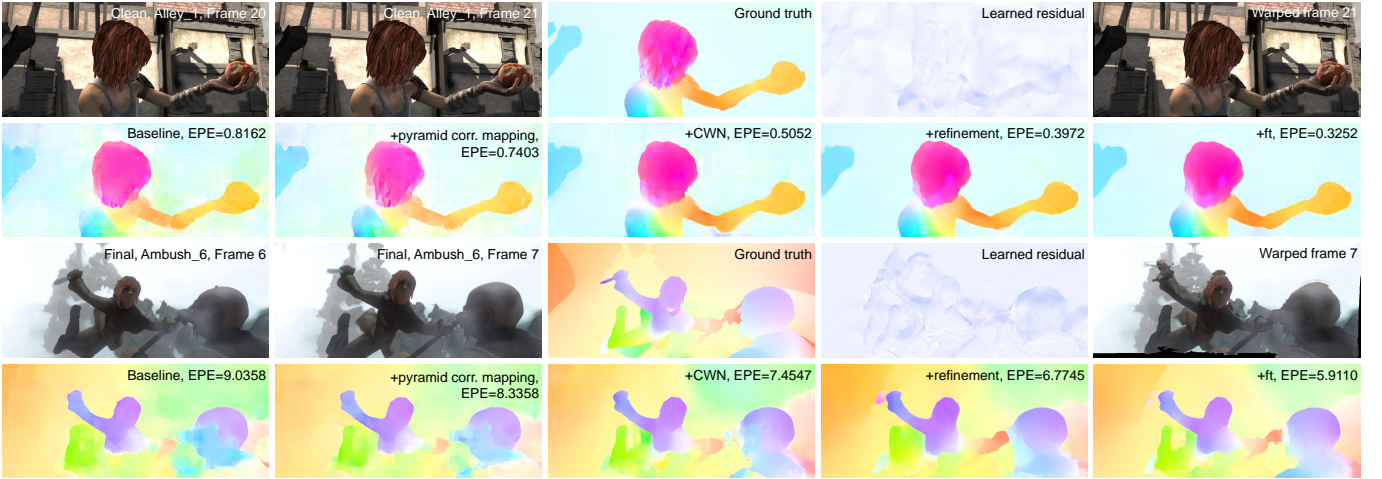}
		%\vspace{-6mm}
		\caption{Results on Sintel training \emph{Clean} and \emph{Final} passes. Pyramid correlation mapping, CWN module and residual learning all improve the performance. And we indicate the learned residual normalized by 1 and warped $I_2$ at the top stage. (Zoom in for details.)}
        \label{fig:ablation}
	\end{center}
\end{figure*}

\begin{figure*}[h]%[p]
	\begin{center}
		\includegraphics[width=\linewidth]{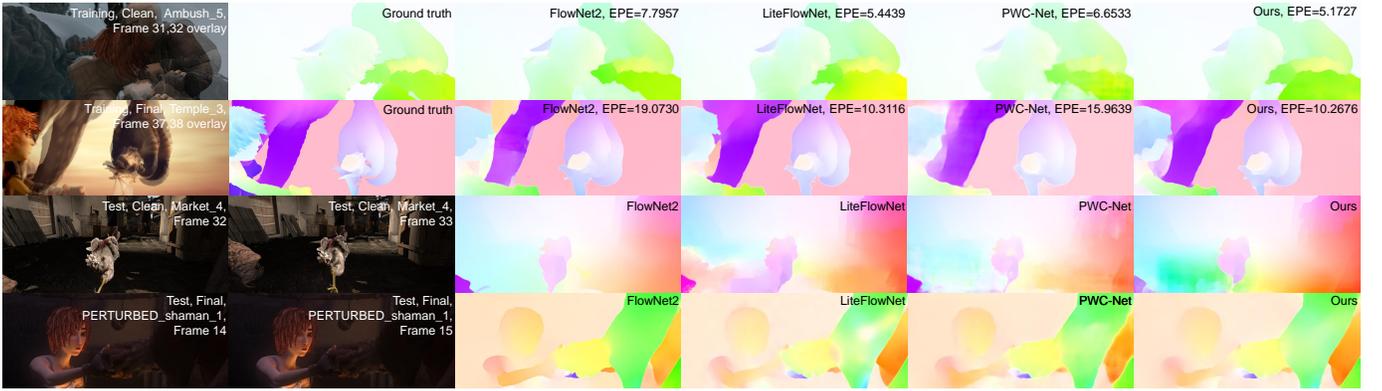}
		%\vspace{-6mm}
		\caption{Examples of predicted optical flow from different methods on Sintel training and test sets for \emph{Clean} and \emph{final} passes. Our method achieves the better performance and preserves the details with fewer artifacts.  (Zoom in for details.)}
        \label{fig:compare}
	\end{center}
%\vspace{-7mm}
\end{figure*}

\begin{table}[t]
 \caption{Comparison of different terms of loss function in Eq. \ref{eq:loss} on Sintel training \emph{Clean} and \emph{Final} passes. The network is trained on FlyingChairs and fine-tuned on FlyingThings3D. ``S'' ``U'' and ``R'' denote Supervised, Unsupervised and Regularization terms in training loss, respectively. }
 \begin{center}
  \label{tab:loss}
   \fontsize{9pt}{11pt}\selectfont\centering
   \begin{tabular}{|l|p{0.8cm}<{\centering}|p{0.85cm}<{\centering}|p{0.85cm}<{\centering}|p{1cm}<{\centering}|} %{|p{2cm}|p{1.3cm}|p{1.5cm}|p{1.9cm}|} %{|l|c|c|c|}
     \hline
     Dataset & S & S+U & S+R & S+U+R \\
     \hline\hline
     Sintel training \emph{Clean} & 2.44 & 2.33 & 2.38 & \textbf{2.24} \\
     Sintel training \emph{Final} & 4.13 & 3.88 & 3.95 & \textbf{3.51} \\
     \hline
   \end{tabular}
 \end{center}
%\vspace{-6mm}
\end{table}

\begin{table}[h]%[t]
%\vspace{-2mm}
 \caption{Ablation study of our component choices of the network. Average end-point error Results of our FPCR-Net with different components of pyramid correlation mapping, CWN module and early up-sampling residual learning branch on Sintel training \emph{Clean} and \emph{Final} passes.}
 \begin{center}
  \label{tab:component}
   \fontsize{9pt}{11pt}\selectfont\centering
   \begin{tabular}{|l|c|c|c|c|c|c|} %{|p{2cm}|p{1.3cm}|p{1.5cm}|p{1.9cm}|} %{|l|c|c|c|}
     \hline
     Baseline & $\surd$ & $\surd$ & $\surd$ & $\surd$ & $\surd$ \\
     Pyramid corr. & $\times$ & $\surd$ & $\surd$ & $\surd$ & $\surd$ \\
     CWN module & $\times$ & $\times$ & $\surd$ & $\surd$ & $\surd$ \\
     Residual reconstruction & $\times$ & $\times$ & $\times$ & $\surd$ & $\surd$ \\
     Finetune & $\times$ & $\times$ & $\times$ & $\times$ & $\surd$ \\
     \hline\hline
     Sintel training \emph{Clean} & 3.11 & 2.78 & 2.67 & \textbf{2.24} & \textbf{(1.58)} \\
     Sintel training \emph{Final} & 4.57 & 3.82 & 3.71 & \textbf{3.51} & \textbf{(1.97)} \\
     \hline
   \end{tabular}
 \end{center}
%\vspace{-6mm}
\end{table}

\begin{table}[h]

\caption{AEE of different methods on the Sintel dataset. The ``-ft'' suffix denotes the fine-tuned networks using the target dataset. The values in parentheses are the results of the networks on the data they were trained on, and hence are not directly comparable to the others.}
%\vspace{-2.5mm}
 %\fontsize{9pt}{10pt}\selectfont\centering
   \begin{center}
   \label{tab:statoftheart}
   %\begin{threeparttable}
   %\tiny %%scriptsize
   \fontsize{9pt}{10.5pt}\selectfont\centering
     %\begin{tabular}{|l|p{0.4cm}<{\centering}p{0.4cm}<{\centering}|p{0.4cm}<{\centering}p{0.4cm}<{\centering}|}
     \begin{tabular}{|p{3.8cm}|cc|cc|}
       \hline
       Method & \multicolumn{2}{|c|}{Sintel clean} & \multicolumn{2}{|c|}{Sintel final} \\
        ~ & train & test & train & test \\
       \hline\hline
       LDOF~\cite{brox2010large} & 4.64 & 7.56 & 5.96 & 9.12 \\
        DeepFlow~\cite{Weinzaepfel2013DeepFlow} & 2.66 & 5.38 & 3.57 & 7.21 \\
        PCA-Layers~\cite{wulff2015efficient} & 3.22 & 5.73 & 4.52 & 7.89 \\
        EpicFlow~\cite{revaud2015epicflow} & 2.27 & 4.12 & 3.56 & 6.29  \\
        FlowFields~\cite{bailer2015flow} & \textbf{1.86} & 3.75 & \textbf{3.06} & 5.81 \\
        Full Flow~\cite{Chen2016Full} & --- & \textbf{2.71} & 3.60 & 5.90 \\
        %\hline
        Deep DiscreteFlow~\cite{guney2016deep} & --- & 3.86 & --- & 5.73 \\
        Patch Matching~\cite{bailer2017cnn} & --- & 3.78 & --- & 5.36 \\
        DC Flow~\cite{Xu_2017_CVPR} & --- & --- & --- & \textbf{5.12} \\
        \hline
        FlowNetS~\cite{dosovitskiy2015flownet} & 4.50 & 7.42 & 5.45 & 8.43  \\
        FlowNetS-ft~\cite{dosovitskiy2015flownet} & (3.66) & 6.96 & (4.44) & 7.76  \\
        FlowNetC~\cite{dosovitskiy2015flownet} & 4.31 & 7.28 & 5.87 & 8.81  \\
        FlowNetC-ft~\cite{dosovitskiy2015flownet} & (3.78) & 6.85 & (5.28) & 8.51 \\
        FlowNet2~\cite{Ilg_2017_CVPR} & \textbf{2.02} & \textbf{3.96} & 3.54 & 6.02  \\
        FlowNet2-ft~\cite{Ilg_2017_CVPR} & (1.45) & 4.16 & (2.19) & 5.74 \\
        %\hline
        SPyNet~\cite{ranjan2017optical} & 4.12 & 6.69 & 5.57 & 8.43 \\
        SPyNet-ft~\cite{ranjan2017optical} & (3.17) & 6.64 & (4.32) & 8.36 \\
        PWC-Net~\cite{Sun_2018_CVPR} & 2.55 & --- & 3.93 & ---  \\
        PWC-Net-ft~\cite{Sun_2018_CVPR} & (2.02) & 4.39 & (2.08) & 5.04  \\
        LiteFlowNet~\cite{hui2018liteflownet} & 2.48 & --- & 4.04 & --- \\
        LiteFlowNet-ft~\cite{hui2018liteflownet} & (1.64) & 4.86 & (2.23) & 6.09  \\
       %\hline
       FPCR-Net (Ours)  & 2.24 & --- & \textbf{3.51} & --- \\
       FPCR-Net-ft (Ours)  & (1.58) & 4.07 & (1.97) & \textbf{4.94} \\
       \hline
     \end{tabular}
     %\begin{tablenotes}
      %\footnotesize
      %\item[*] Finetuned from Kinetics dataset.
    %\end{tablenotes}
  %\end{threeparttable}
   \end{center}
%\vspace{-6mm}
\end{table}
\noindent\textbf{Effectiveness of CWN Module.} The CWN module fuses correlation features from pyramid correlation mapping and the coarse flow from the last stage efficiently for accurate flow estimation. We show the performance of architectures with different components and correlation features from pyramid correlation mapping in Table \ref{tab:cwn}. The entire CWN module with pyramid correlation mapping get the better performance and each component makes contributions to the network with more correlation details, less warping error and accurate relative values.

%Aggregation of multi-level correlated cost volume provides the opportunity to jointly explore the motion details at the video level. We show the performance of our scheme in comparison with the baseline scheme on the Sintel training \emph{Clean} dataset in Table \ref{tab:}.
%
%In Table \ref{tab:corr}, ``Single corr.'', ``Pyramid corr.'', and ``Pyramid corr. mapping'' correspond to the three schemes in Figure \ref{fig:corr}, respectively. We can see that ``Pyramid corr. mapping'' scheme achieves 0.20 and 0.32 decrease in AEE measured in Sintel training \emph{clean} trained by the FlyingChairs dataset and fine-tuned by the FlyingThings3D dataset.

\noindent\textbf{Comparisons on Residual Learning Branch Designs.} We have designed a residual learning branch for finer optical flow reconstruction. The purpose of this branch consists of a transposed convolution layer and convolution units in Figure \ref{fig:refine_conv} to explore motion details and occlusion compensation between frames to learn fine-grained residual representation. We have tried three network structures with different locations of up-sampling shown as Figure \ref{fig:refinement}. The numbers of convolution unit output channels of the entire branch are 64, 64, 128 and 256. We show the results of these designs in Table \ref{tab:refine} with experiments conducted on the Sintel training \emph{Clean} pass. We can see that the early up-sampling scheme achieves better performance than the others with detail-preserving flow field learned from large-scale features.

\noindent\textbf{Comparisons on Different Loss Term Combination.} We have designed the loss function for self-learning and regularization. We try to compare the loss of different components in Table \ref{tab:loss}. The all-term utilized loss indicates the better performance on Sintel training \emph{Clean} and \emph{Final} passes. The results indicate that it is effective to utilize different types of loss term for parameter optimization.

We explore the contribution of each component option by calculating the AEE with some of the components enabled or disabled in Table \ref{tab:component}.  Figure \ref{fig:ablation} illustrates the component-accumulated examples of flow fields on the Sintel dataset. We can see that the small-magnitude artifacts are restrained and the smaller AEE is achieved with the components accumulated. In addition, we extract the features of warped image $\tilde I_2$ for unsupervised loss term and learned residual details on the last stage. By involving these useful items, the network addresses accurate optical flow with detail-preserving.

\subsection{Comparison with State-of-the-art Methods}

We compare our proposed scheme with state-of-the-art approaches for video action recognition in Table~\ref{tab:statoftheart}. We evaluate the performance on the Sintel dataset. For both datasets, we use the provided evaluation protocol and report the AEE. We can see that our scheme achieves the best performance, with 4.07 and 4.94 of AEE on the \emph{Clean} and \emph{Final} passes of Sintel dataset, especially on the \emph{Final} pass with a significant performance, with improvement by 0.80, 1.15 and 0.10 in terms of AEE against FlowNet2, LiteFlowNet and PWC-Net.%and the KITTI

Figure \ref{fig:compare} demonstrates our results compared with some competing baseline methods --- FlowNet2, LiteFlowNet, and PWC-Net on the Sintel dataset. We can see that the network gets the better performance and finer details are well preserved with fewer artifacts of our method. However, the smaller or slimmer objects are the challenges for the network of which edges and patches are not preserved effectively, and we will perform further study on the more challenging cases in the future.

\section{Conclusion}

To model the motion details in videos for accurate optical flow estimation, we propose a pyramid correlation mapping and residual reconstruction framework --- FPCR-Net, to enable the joint analysis of pyramid cost volume and the refinement by stages. The pyramid correlation mapping module takes advantage of the multi-scale correlations of both global and local patches by aggregating features of different scales, while the residual reconstruction module aims to reconstruct the sub-band high-frequency residuals of finer optical flow in each stage. Experiment results show that the proposed scheme achieves the state-of-the-art performance, with improvement by 0.80, 1.15 and 0.10 in terms of AEE against FlowNet2, LiteFlowNet and PWC-Net on the \emph{Final} pass of Sintel dataset, respectively.

\ifCLASSOPTIONcaptionsoff
  \newpage
\fi

\bibliographystyle{IEEEtran}

\bibliography{reference}

%\vfill

% that's all folks
\end{document}